\documentclass[nonacm]{acmart}
\usepackage{balance}
\usepackage{natbib}
\usepackage{hwemoji}
\usepackage{algorithm}
\usepackage{algorithmic}
\usepackage{graphicx}

\AtBeginDocument{%
  \providecommand\BibTeX{{%
    \normalfont B\kern-0.5em{\scshape i\kern-0.25em b}\kern-0.8em\TeX}}}

\usepackage{amsmath}
\usepackage{xcolor}
\usepackage[framemethod=tikz]{mdframed}
\usetikzlibrary{shadows}

\definecolor{mygray}{HTML}{F5F5F5}

\begin{document}

\title[Limited Effectiveness of LLM-based Data Augmentation]{Limited Effectiveness of LLM-based Data Augmentation for COVID-19 Misinformation Stance Detection}

\author{Eun Cheol Choi}
\orcid{0000-0003-0861-1343}
\email{euncheol@usc.edu}

\affiliation{%
  \institution{University of Southern California}
  \streetaddress{3502 Watt Way}
  \city{Los Angeles}
  \state{California}
  \country{USA}
  \postcode{90089-0281}
}

\author{Ashwin Balasubramanian}
\orcid{0009-0006-6742-0933}
\email{ashwinba@usc.edu}

\affiliation{%
  \institution{University of Southern California}
  \streetaddress{3502 Watt Way}
  \city{Los Angeles}
  \state{California}
  \country{USA}
  \postcode{90089-0281}
}

\author{Jinhu Qi}
\orcid{0009-0006-5544-4786}
\email{qijinhu1218@gmail.com}

\affiliation{%
  \institution{University of Southern California}
  \streetaddress{3502 Watt Way}
  \city{Los Angeles}
  \state{California}
  \country{USA}
  \postcode{90089-0281}
}

\author{Emilio Ferrara}
\orcid{0000-0002-1942-2831}
\email{emiliofe@usc.edu}

\affiliation{%
  \institution{University of Southern California}
  \streetaddress{3502 Watt Way}
  \city{Los Angeles}
  \state{California}
  \country{USA}
  \postcode{90089-0281}
}

\renewcommand{\shortauthors}{Choi et al.}

\begin{abstract}
Misinformation surrounding emerging outbreaks poses a serious societal threat, making robust countermeasures essential. One promising approach is stance detection (SD), which identifies whether social media posts support or oppose misleading claims. In this work, we finetune classifiers on COVID-19 misinformation SD datasets consisting of claims and corresponding tweets. Specifically, we test controllable misinformation generation (CMG) using large language models (LLMs) as a method for data augmentation. While CMG demonstrates the potential for expanding training datasets, our experiments reveal that performance gains over traditional augmentation methods are often minimal and inconsistent---primarily due to built-in safeguards within LLMs. We release our code and datasets to facilitate further research on misinformation detection and generation.
\end{abstract}




\begin{teaserfigure}
  \centering
  \includegraphics[width=.8\textwidth]{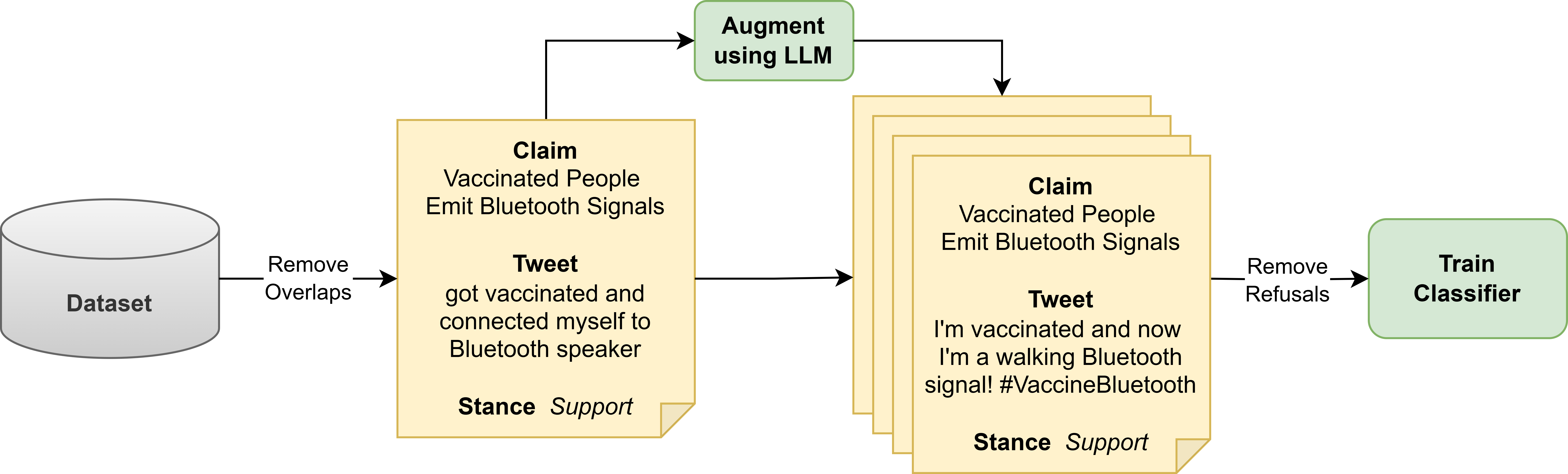}
  \caption{Workflow of this paper with illustrative examples \cite{malashenko2023bluetooth}.}
  \Description{An overview of data augmentation process on COVID-19 stance detection datasets.}
  \label{fig:teaser}
\end{teaserfigure}

\maketitle

\section{Introduction}
While misinformation on social media poses a pressing concern, users can help correct it by refuting false claims \cite{ma2023characterizing}. Research suggests that corrections can reduce belief in misinformation \cite{walter2020fact}, and repetition does not typically backfire \cite{swire2020searching}. Hence, efforts should concentrate on discouraging misinformation and amplifying refutative posts, requiring robust detection methods to distinguish opposing posts from supporting ones.

We focus on a stance detection (SD) task for pairs of COVID-19 false claims and social media posts, building upon prior research \cite{choi2024fact}. SD is crucial because it allows models to detect rebuttals or endorsements of misleading content, enabling timely interventions. Despite the abundance of COVID-19 datasets today, the early pandemic lacked sufficient labeled data, making it difficult to train robust classifiers. To address this challenge, we simulate data-scarce scenarios and evaluate various data augmentation techniques. In particular, we assess the effectiveness of controllable misinformation generation (CMG) \cite{chen2023can} in enhancing model performance.

Inspired by prior studies, we experiment with CMG to generate diverse misinformation-related tweets using large language models (LLMs) for training data augmentation. Ideally, CMG can simulate endless variations of social media posts related to misinformation \cite{nakov2021automated}, producing datasets that improve classifier generalization to newly emerging misinformation. However, CMG provides only minimal advantages over more cost-effective data augmentation methods, particularly in low-resource scenarios with limited real-life data. We discuss these issues and attribute them in part to the safeguards implemented in chat models. The code and datasets used in this study are available at \textcolor{blue!60!black}{\url{https://doi.org/10.5281/zenodo.14885088}}.

\section{Datasets} \label{sec:dataset}

Our study uses datasets for natural language inference (NLI) and stance detection (SD) tasks that pair COVID-19 false claims with tweets. Both tasks analyze the relationship between a text and a target: NLI assigns entailment, contradiction, or neutral labels, while SD identifies support, opposition, or neither \cite{khiabani2024cross}. By aligning NLI labels with SD labels, we leverage large-scale NLI datasets to improve SD performance \cite{khiabani2024cross, zhao2023ez}.

We first utilized the dataset from \textit{FACT-GPT} \cite{choi2024fact}, which pairs COVID-19 false claims with tweets using 3-class labels. We expanded this dataset by annotating 900 additional claim--tweet pairs. We also incorporated two additional COVID-19 SD datasets, \textit{Covid19-Lies} \cite{hossain2020covidlies} and \textit{COVMis-Stance} \cite{hou2022covmis}, from which 52.0\% of tweets were retrieved via the Twitter API. Unretrieved tweets were deleted or belonged to protected accounts.

We removed trivial pairs with 100\% word overlap after lemmatization to ensure data quality. To mitigate data leakage, we grouped pairs with over 90\% lexical overlap into the same split (training, development, or test).

\section{Experiments}

We employed a two-step finetuning process to leverage broader NLI capabilities before focusing on COVID-19-specific stance detection. We first established a pretrained model using a general NLI Cross-encoder \cite{reimers2019sentence}, built on a \texttt{deberta-v3-base} model \cite{he2021deberta}. This model was initially tuned using \textit{SNLI} \cite{bowman-etal-2015-large} and \textit{MNLI} \cite{N18-1101} for general NLI ability, as well as \textit{FEVER-NLI} \cite{nie2019combining} for fact-checking capabilities. Next, we used various strategies to augment the COVID-19 SD data for finetuning purposes, increasing it by four times its original size. For comparison, we also finetuned a model solely on real-life data. We evaluated the models' performance using ground-truth annotations in the test set. We performed all local inference and training processes using a single A40 GPU, ensuring consistent computational resources throughout the experiment.

The previous works and datasets consistently have reported the imbalanced nature of the task \cite{pado2022textual, hardalov2021survey}. We opted not to employ under- or oversampling, which are common interventions for class imbalance. Instead, we focused on data augmentation to boost the size of the training set while maintaining the original class distribution. By maintaining the original class distribution during augmentation, we could ensure that the model learns from a dataset that reflects the real-world distribution of the task.

\subsection{Augmentation Strategies} \label{sec:augstrat}
We simulated low-resource scenarios to evaluate the effectiveness of various data augmentation techniques. This approach is especially relevant for emerging events where labeled data is scarce, such as disease outbreaks. Specifically, we examined performance across datasets containing 100, 200, 500, 1,000, 2,000, and 5,000 claim–tweet pairs. 

We compared several augmentation strategies:

\begin{itemize}
    \item Controllable Misinformation Generation (CMG) \cite{chen2023can}: Generating misinformation posts with LLMs.
    \item Easy Data Augmentation (EDA) \cite{wei2019eda}: Randomly applying simple text operations like synonym replacement, word swapping, etc.
    \item An Easier Data Augmentation (AEDA) \cite{karimi2021aeda}: Randomly applying punctuation insertions.
    \item Backtranslation \cite{ma2019nlpaug}: Translating text to intermediate languages and back, creating paraphrased versions of the original text.

\end{itemize}

In this study, CMG is defined as:
\begin{itemize}
    \item \textbf{Given:} The training set of false claim (\( C_i \)), social media post (\( P_i \)), and label (\( L_i \)) triplets:
    \[
    T = \{(C_1, P_1, L_1), (C_2, P_2, L_2), \ldots, (C_n, P_n, L_n)\}
    \]

    \item \textbf{Objective:} Generate a new \( P'_m \) for an existing \( (C_m, L_m) \).

    \item \textbf{Approach:}  
    \[
    P'_m = \text{Generate}(C_m, L_m \mid S), \quad S \subset T
    \]
    where \text{Generate} is a forward pass through an LLM, and \( S \) consists of example(s) provided as shot(s).
\end{itemize}

For data augmentation ($C_m$, $P'_m$, $L_m$) is appended to \textit{T}. However, to ensure high-quality augmentation, we filtered out any CMG-generated samples with a rejection score greater than 0.9, as determined by the \texttt{DistilRoBERTa}-based rejection classifier \cite{distilroberta-base-rejection-v1}. We report LLM generation refusal rates in Section~\ref{sec:refusal}. We experimented with \texttt{GPT-4o}, \texttt{GPT-3.5-Turbo}, \texttt{LLaMa-3.1-8B-Instruct}, and \texttt{Qwen2.5-7B-Chat} for the CMG approach.

\begin{table}[t]
\centering
\caption{Description of annotated datasets of COVID-19-related claim-tweet pairs leveraged in this paper. Each label shows whether the given tweet supports/opposes/is unrelated to the claim.}
\begin{tabular}{lcccc}
\toprule
\textbf{Name/Doc} & \textbf{Support} & \textbf{Oppose} & \textbf{Neither} & \textbf{Total}\\
\midrule
% Add your data rows here
FACT-GPT \cite{choi2024fact} & 682 & 111 & 403 & 1196 \\
+ Extension & 197 & 53 & 650 & 900 \\
Covid19-Lies \cite{hossain2020covidlies} & 103 & 63 & 2551& 2717 \\
COVMis-Stance \cite{hou2022covmis} & 944 & 812 & 250 & 2006\\

\bottomrule
\end{tabular}
\end{table}

For EDA and AEDA, we implemented the techniques described in their respective repositories. Backtranslation was performed using four languages: Italian, German, Chinese, and Russian, using NLP Augmentation library \cite{ma2019nlpaug} and Opus-mt models \cite{tiedemann2020opus}. 

Our experimental configurations allow us to compare traditional augmentation techniques and LLM-based approaches across various data availability scenarios.

\begin{figure}[t]
    \centering
    \begin{tikzpicture}[font=\small]
        \node[draw, fill=mygray, rounded corners, drop shadow={fill=black!30, shadow xshift=1pt, shadow yshift=-3pt, opacity=0.5}, inner sep=10pt] {
            \begin{minipage}{0.4\textwidth}
                \begin{tabular}{l p{0.75\textwidth}}
                    \texttt{System} & Generate TWEET so that if TWEET is true, then CLAIM \{\underline{is also true}, is false, cannot be said to be neither true nor false\}. Try your best to mimic the styles of example TWEETs. Respond with only a single TWEET. \\
                    \texttt{Input} & {\{\textit{example pairs}\}} \\
                     & CLAIM: 2019 novel Coronavirus can be cured by one bowl of freshly boiled garlic \\
                     & TWEET: \\
                    \texttt{Output} & Garlic soup is the secret weapon against COVID-19? Just one bowl and you're cured! 🧄💪 \#GarlicPower \#NaturalRemedy \\
		      \end{tabular}
            \end{minipage}
            };
    \end{tikzpicture}
    \caption{Example CMG prompt conditioned on few-shots, a false claim \cite{factly2020garlic}, and a task description \cite{choi2024fact}.}
    \label{fig:gen_prompt}
\end{figure}

\begin{figure}[t]
    \centering
    \includegraphics[width=.6\textwidth]{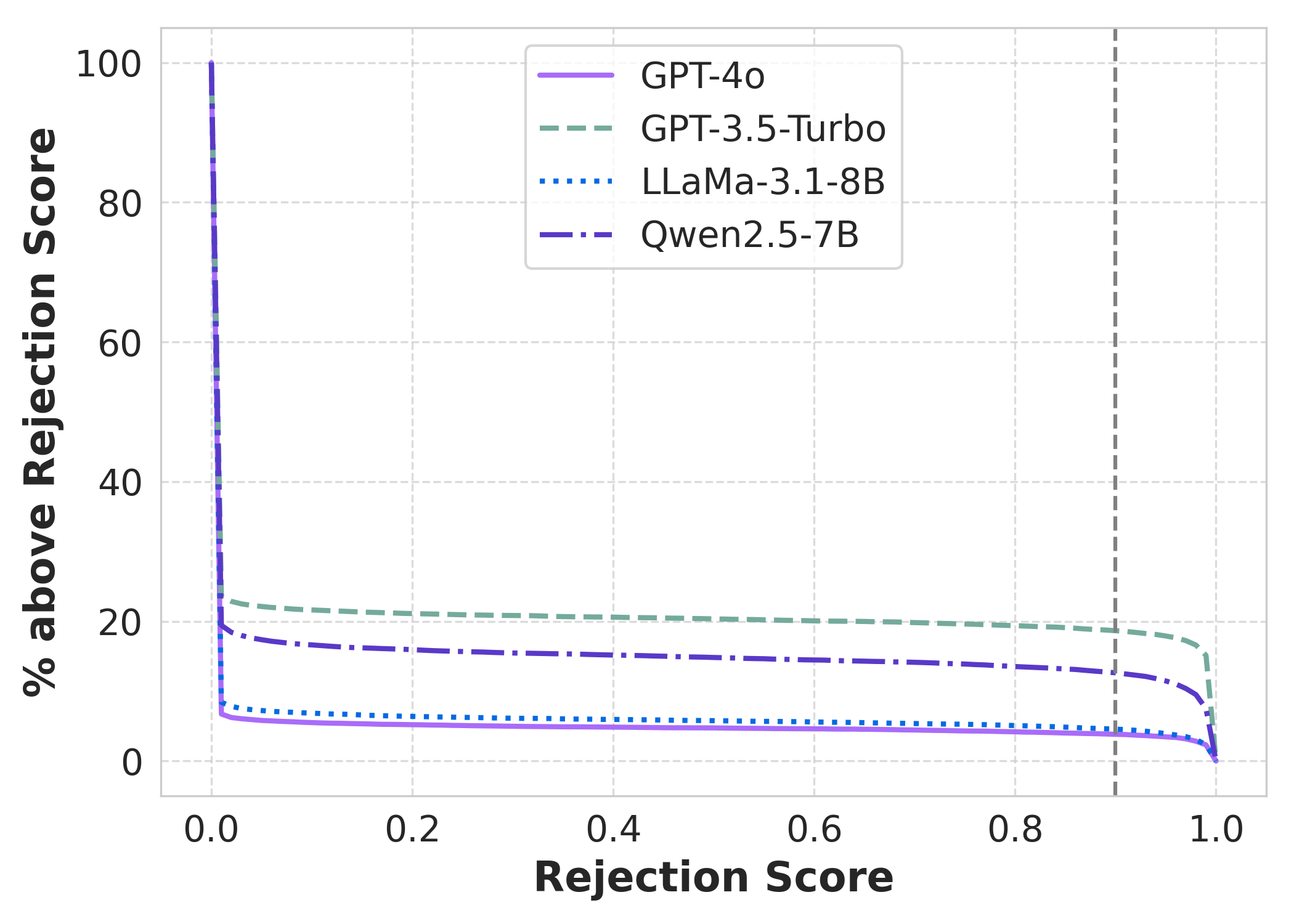}
    \caption{Refusal scores of different LLMs' outputs on CMG task, measured with a rejection classifier \cite{distilroberta-base-rejection-v1}.}
    \label{fig:rejection_rate}
\end{figure}

\subsection{Results}

\subsubsection{CMG Refusal} \label{sec:refusal}

To assess the reliability of different LLMs in generating misinformation-related content, we analyzed their refusal rates using the rejection classifier \cite{distilroberta-base-rejection-v1}. Figure \ref{fig:rejection_rate} illustrates various models' overall refusal score distributions. We set the threshold to 0.9, accounting for the bimodal nature of the score distribution.

Our analysis reveals significant differences in refusal rates among the tested models. \texttt{GPT-4o} demonstrates the lowest refusal rate, with 3.8\% of its generated content classified as refusal at a threshold of 0.9. \texttt{GPT-3.5-Turbo}, \texttt{LLaMa-3.1-8B-Instruct}, and \texttt{Qwen2.5-7B-Chat} show refusal rates of 18.7\%, 4.6\%, and 12.6\%, respectively. These differences suggest that some models are more heavily moderated than others, likely due to stricter safeguards implemented during training or finetuning processes.

These findings highlight the challenges in using LLMs for CMG in research settings. Heavily moderated models may significantly limit the diversity and representativeness of generated content. This variability underscores the importance of considering the level of safeguards when using LLMs for misinformation generation research.

\subsubsection{Performance and Limitations of CMG Data Augmentation} \label{sec:performance_classifier}

\begin{table}
\small
  \caption{Performance of classifiers. Original training set \textit{N}=5000, augmented by adding four times the original size. Average of five finetuning runs.}
  \label{tab:performance}
  \begin{tabular}{@{}l@{\hspace{0.5em}}c@{\hspace{0.5em}}c@{\hspace{0.5em}}c@{\hspace{0.5em}}c@{\hspace{0.5em}}c@{}}
    \toprule
    \textbf{Training Strategy} & \textbf{$F1_{Macro}$} & \textbf{$Accuracy$} & \textbf{$F1_{Sup}$} & \textbf{$F1_{Opp}$} & \textbf{$F1_{Nei}$} \\
    \midrule
    \textbf{Not Augmented}              & &  & & & \\
    \texttt{Pretrained}        & .403 & .447 & .402 & .266 & .541 \\
    \texttt{Pre-/Finetuned}    & .773 & .796 & .733 & .718 & .869 \\
    \midrule        
    \textbf{Augmented} & & & & &\\
    \texttt{EDA}               & .783 & .807 & .763 & .724 & .861 \\
    \texttt{AEDA}              & .722 & .752 & .693 & .659 & .815 \\
    \texttt{Backtranslation}   & .789 & .810 & .758 & \textbf{.745} & .863 \\
    \midrule
    \textbf{Augmented, CMG (3-shot)}&  &  &  &  &  \\
    \texttt{GPT-4o}           & \textbf{.793} & \textbf{.818} & \textbf{.764} & .733 & \textbf{.882} \\
    \texttt{GPT-3.5-Turbo}    & .692 & .713 & .686 & .627 & .762 \\
    \texttt{LLaMa-3.1-8B-Instruct} & .787 & .816 & .794 & .693 & .873 \\
    \texttt{Qwen2.5-7B-Chat}  & .765 & .801 & .775 & .651 & .869 \\
    \bottomrule
  \end{tabular}
\end{table}

\begin{figure}[t]
    \centering
    \includegraphics[width=.6\textwidth]{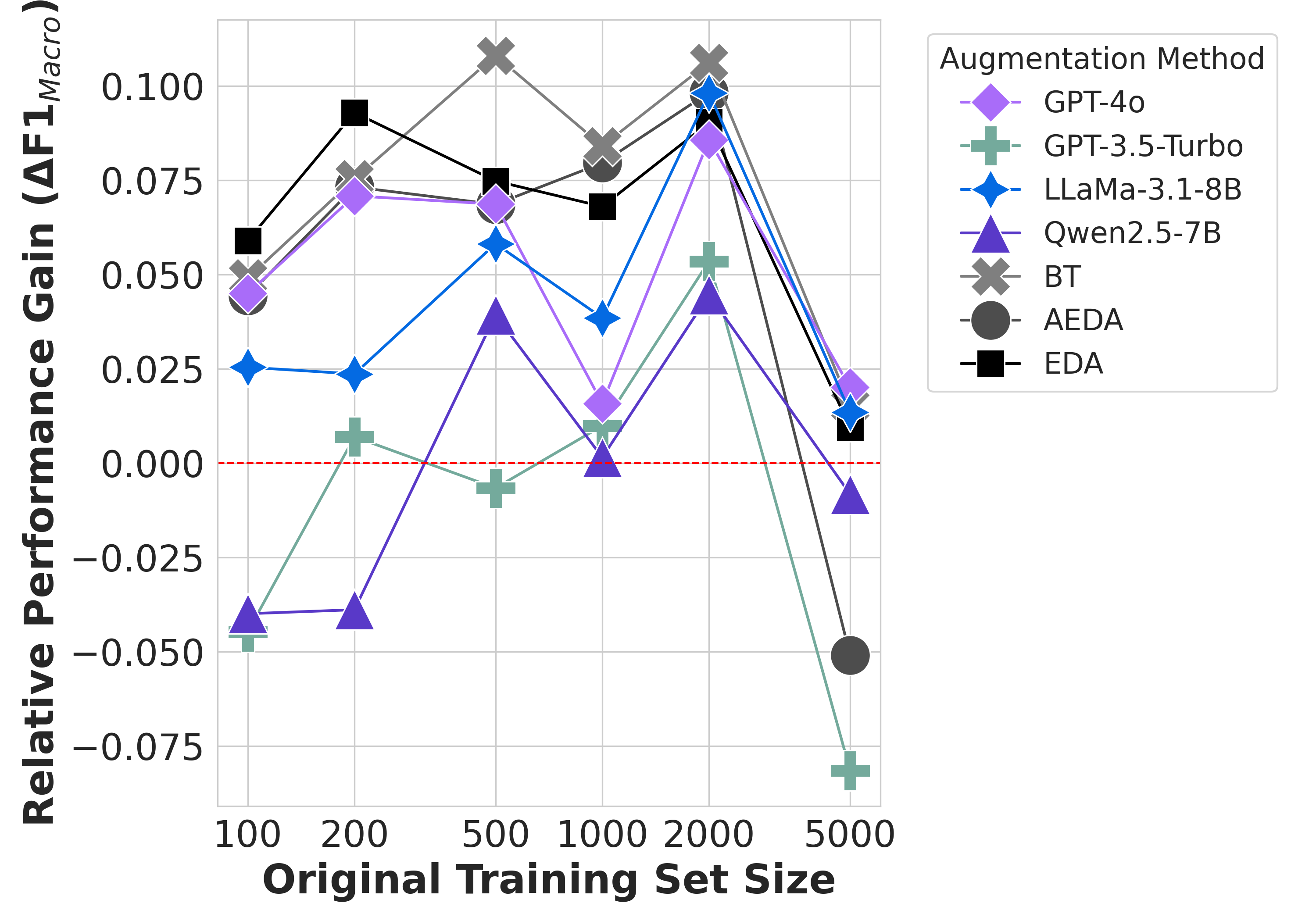}
    \caption{Performance trend compared to a model finetuned on the unaugmented dataset (red horizontal dotted line).}
    \label{fig:performance_trend}
\end{figure}

Our experiments highlight the limitations of CMG as an augmentation strategy. Table \ref{tab:performance} shows that LLM-based CMG yields only marginal improvements in $F1_{\text{Macro}}$ and accuracy compared to more straightforward methods such as backtranslation, EDA, and AEDA. Compared to the finetuned baseline model without data augmentation, CMG produced little to no performance gains depending on the LLM used. Additionally, as shown in Figure \ref{fig:performance_trend}, performance gains from CMG remain inconsistent across varying data sizes. In many scenarios, less resource-intensive data augmentation methods outperform CMG pipelines, highlighting challenges in current LLM-based approaches for data augmentation.

Two primary factors contribute to these limitations:

\begin{itemize}
    \item \textbf{Refusals:} LLMs often refuse to generate certain types of content, limiting the data points available for augmentation.
\end{itemize}

For example, LLMs often reject requests for misinformation-related content by responding with messages such as, ``Sorry, I cannot generate content that promotes misinformation or conspiracy theories. If you have any other topic in mind, feel free to ask!". Although these responses were effectively filtered out with a rejection classifier (Section~\ref{sec:augstrat}), this process reduced the number of training samples.

\begin{itemize}
    \item \textbf{Task flipping:} Even when LLMs generate content, they may inadvertently alter the nature of the task, potentially introducing noise into our dataset.
\end{itemize}

For instance, when we experimented with CMG using \texttt{GPT-4o}, asking it four times to generate tweets supporting the false claim ``2019 novel Coronavirus can be cured by one bowl of freshly boiled garlic'' \cite{factly2020garlic}, it correctly generated supporting tweets 3 times, such as ``Garlic soup is the secret weapon against COVID-19? Just one bowl and you're cured! 🧄💪 \#GarlicPower \#NaturalRemedy''. However, it also incorrectly generated: ``Garlic is amazing for flavor, but let us not kid ourselves—it is not a cure for COVID-19. Stick to science, not soup recipes, when it comes to your health! 🧄💉 \#COVID19 \#TrustScience \#StaySafe''. 

These two kinds of behaviors, refusals, and task flipping, are most likely to be byproducts of aligning LLMs to establish safety guards against generating harmful and deceptive content \cite{chen2023can}.

\section{Conclusions}

This study compares different approaches for augmenting misinformation SD datasets. Our findings reveal that data augmentation via CMG with generative LLMs offers minimal to no advantages over traditional augmentation techniques. We hypothesize that this limitation is partly due to the safeguards implemented in customer-facing chat and instruction-tuned models widely used by the general audience. However, since our study focused on COVID-19 misinformation, its findings may not fully generalize to other domains.

\paragraph{Implications for future research} The constraints of current LLMs in generating misinformation for research purposes underscore the need for more flexible models in academic settings. Future efforts should prioritize designing LLMs with adjustable safeguards, enabling controlled misinformation generation to support detection and analysis. Expanding this work to address misinformation in diverse domains beyond COVID-19, such as political disinformation or health myths, would provide valuable insights into the wider applicability of the findings. Finally, developing more sophisticated methods for mitigating AI-generated misinformation content remains a crucial area for future work.

\section{Related Work}

SD and NLI are central to misinformation detection, typically employing neural networks and pretrained models \cite{zeng2017neural, dulhanty2019taking}. Integrating and SD and NLI features has improved fake news detection and efforts to combat misinformation \cite{hardalov2021survey}.

LLMs offer opportunities for detecting misinformation and augmenting training data \cite{satapara2024fighting, chen2023combating}. However, caution is necessary when working with sensitive datasets like fake news or toxicity, since LLMs may refuse to process such data \cite{piedboeuf2023chatgpt}. Because LLMs can be exploited to generate deceptive content at scale \cite{chen2023can}, continuous research into misinformation detection and generation remains essential.

\balance
\bibliographystyle{abbrvnat}

\bibliography{www25augmis}

\end{document}